\documentclass{article}
\usepackage{spconf,amsmath}
\usepackage{graphicx}
\usepackage{caption}
\usepackage{subcaption}
\usepackage{color}

\title{Explainability in CNN Models by Means of Z-scores}
\name{David Malmgren-Hansen, Allan Aasbjerg Nielsen and Leif Toudal Pedersen}
\address{Technical University of Denmark}

\begin{document}
%
\maketitle
\begin{abstract}
This paper explores the similarities of output layers in Neural Networks (NNs) with logistic regression to explain importance of inputs by Z-scores. The network analyzed, a network for fusion of Synthetic Aperture Radar (SAR) and Microwave Radiometry (MWR) data, is applied to prediction of arctic sea ice. With the analysis the importance of MWR relative to SAR is found to favor MWR components. Further, as the model represents image features at different scales, the relative importance of these are as well analyzed. The suggested methodology offers a simple and easy framework for analyzing output layer components and can reduce the number of components for further analysis with e.g. common NN visualization methods.
\end{abstract}
\begin{keywords}
Convolutional Neural Networks, Z-scores, Logistic Regression, Explainability
\end{keywords}
\section{Introduction}
\label{sec:intro}

Explainability of non-linear complex machine learning models such as Deep Neural Networks remains a challenging time-consuming tasks with increasing importance. As the use of these models spreads into new areas, explainability becomes important to obtain new knowledge about scientific problems or to understand ethical dilemmas that may arise when trusting the outcome without understanding how it was reached. 

In this work we study how to interpret a Convolutional Neural Network (CNN) designed for Satellite Sensor Fusion. The aim is to exploit the statistical similarities an output layer of a CNN shares with the corresponding linear model from common statistical practices. 
The network in our experiments was designed for prediction of sea ice in the Arctic. Currently, mapping of sea ice is a manual discipline where ice analysts draw polygons on top of Synthetic Aperture Radar (SAR) images and assign these polygons average ice concentration intervals. The potential improvements for automating the time consuming manual process of drawing these polygons, i.e.\ Ice Charts, lie in larger and more frequent coverage, and increased consistency with less manual labour. Automating the process on SAR data alone is challenging. SAR images show patterns related to ice formations but back-scatter intensities can be ambiguous between ice and open water. 
The model in this study tackles the challenges by fusing SAR data from Sentinel-1 with Microwave Radiometer (MWR) data from AMSR2 to exploit the advantages of each instrument. While SAR data has ambiguities it has a very high resolution, where MWR data has good contrast between open water and ice, the resolution is too poor for many applications of sea ice mapping. 

In the last layer of a CNN, several features are concatenated in the image domain before a product-sum of a set of coefficients is passed through a Sigmoid function to form probabilities of ice in a pixel in the image. This process is similar to a logistic regression except for the CNN being optimized by stochastic gradient descent in an end-to-end manner. Convolutional filters that extract image features from the SAR data are learned simultaneously with the coefficients in the mixing layer that combines the information with MWR data. Interpreting these coefficients provides information on how the CNN weights the importance of the inputs, which is exploited in this paper. This methodology is further explained in Section \ref{sec:model}.

\section{Related Work}\label{sec:RW}
Interpretation and visualization of features and filters in CNNs were pioneered by \cite{zeiler2014visualizing}. Since their work on analyzing which pixels in a given image are important for the decision of a CNN, several improvements and new methods have followed \cite{bach2015pixel,Zhou2015OBD,zhou2018interpreting}. One of the biggest remaining challenges with CNNs and Deep Learning is the lack of explainability and interpretation, as adressed in \cite{eldar2017challenges}. In remote sensing less work has been done with model interpretation than in general computer vision, and regarding fusion networks hardly any papers have touched upon this subject. In \cite{benzenati2019generalized}, they mention this problem as to easily be impressed by empirical improvements of Pan Sharpening and multi-spectral fusion networks without any notion to the correctness of the spectral information of the fused product. The authors in \cite{benzenati2019generalized} propose a method to control which information from the Panchromatic image band is used in the fusion network.

As opposed to previous work in the field, this paper focuses on the interpretability of the last layer in a satellite sensor fusion network. In many CNN model architectures the last layer's inputs consist of semantic feature representations learned by the model. By knowing which inputs are important one can deal with fewer nodes to interpret. 

\section{Model and Experiments}\label{sec:model}
The model presented here extracts image features from the SAR images at four different scales by means of an Atrous Spatial Pyramid Pooling, \cite{chen2017rethinking}. In the networks last layer a linear mixing of feature inputs with MWR components allows for a direct interpretation of what information is important in order to predict sea ice.

\begin{figure*}[htb]
    \centering
    \includegraphics[width=0.8\linewidth]{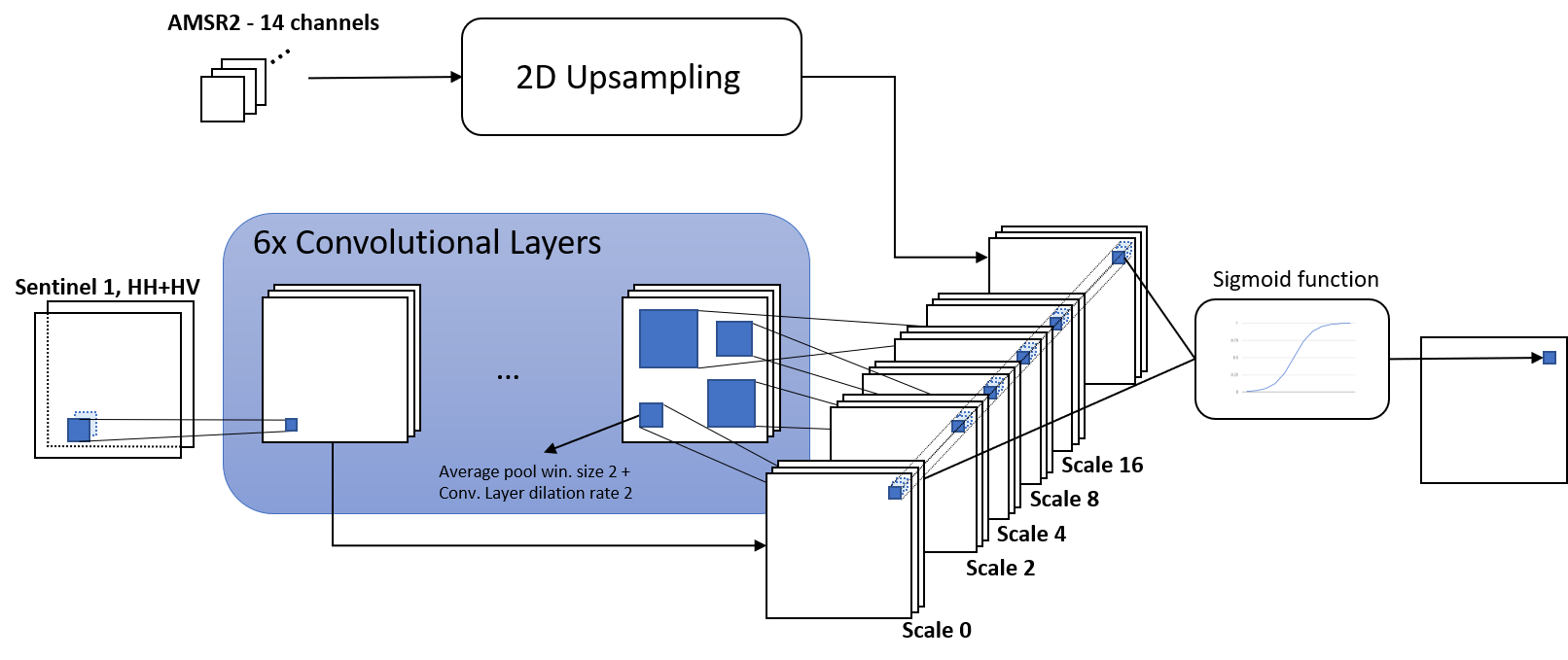}
    \caption{Model architecture of the sensor fusion model analyzed in this work. The different scales are controlled by dilation rates i.e. scale 2 is dilation rate equal 2. Before applying dilation filters the input is smoothed with a average operation in a $dxd$ window with $d$ equal to the dilation rate.}
    \label{fig:model}
\end{figure*}

As the SAR images and MWR data have very different resolutions, the model applies a series of convolutions on the SAR data independently. In the second to last layer a set of different features at different scales are extracted by using different dilation rates in the convolutional filters. These are then concatenated with upsampled AMSR2 brightness temperature measurements in order to perform a pixel-wise linear combination that is passed through a sigmoid function, see Figure \ref{fig:model}.

Due to the resolution difference between AMSR2 and Sentinel-1 data of up to a factor 268,000 (for 6.93GHz ASMR2 band wrt.\ area of footprint), it is computationally unwise to perform equally many convolutions on the two input sources. Therefore, this CNN processes the SAR input separately and merges it with the coarser MWR at a later stage.
The image feature scales are not achieved by sub-sampling schemes but by the use of dilated convolutions with four different dilation rates. One set of image features is extracted by skip-connection from the output of the second convolution layer and is referred to as Scale-0. This scale has also skipped all batch-normalization operations and therefore retains a linear relationship to the original back-scatter values in the Sentinel-1 image. Scales 2-16 have been processed by six convolutional layers pair-wise interleaved with ReLU activations, Batch Normalization and Dropout Regularization. This means the distribution of activation intensities are normalized over batches of data inside the network.

The interesting property of this CNN architecture is that all inputs to the last layer represent different information about the problem at hand. It is relevant to know if Scale-0 is more important to the problem than Scale-16 which might have an insufficient resolution to capture edge information in the ice charts. Another question, is what effect the MWR data has? MWR instruments measure with far coarser resolution but have high contrast between open water and ice, almost independent of wind and ocean current conditions.

As the output size of a convolution layer is an adjustable parameter, one can also study what effect it has on the importance of a feature scale to increase the size. This will be explored by two model variants, one with increased sizes of output nodes in the feature Scales 2-16. Table \ref{tab:1} shows the sizes of the inputs to the last CNN layer for the two model variants studied here. 
\begin{table}\caption{Input size of each input type for the two model variants.}\label{tab:1}
\centering
\begin{tabular}{|l|l|l|}
\hline
\textbf{Inp. Type} & \textbf{Small} & \textbf{Large} \\ \hline
Scale-0 & 14 & 14 \\ \hline
Scale-2 & 14 & 28 \\ \hline
Scale-4 & 14 & 28 \\ \hline
Scale-8 & 14 & 28 \\ \hline
Scale-16 & 14 & 28 \\ \hline
AMSR2 & 14 & 14 \\ \hline
\end{tabular}
\end{table}

Section \ref{sec:res} elaborates on
the effect of the increased size of scale-feature layers in the CNN on the balance between Sentinel-1 and between Sentinel-1 and AMSR2 components.

\section{Statistical Properties of Neural Networks}
\label{sec:pagestyle}

The last layer of a Neural Network trained with categorical cross-entropy, i.e. image classification or per-pixel classification, resembles a logistic regression since it is a linear combination of preceding layer's outputs parsed through a sigmoid function. In our case the last layer processes a concatenation of ASMR2 channels with features extracted from the SAR image. This allows for a straight forward comparison between the importance of the two data sources. As typically done in logistic regression, coefficients can be corrected with input standard deviation to provide Z-scores,
\begin{equation}\label{eq:zscore_org}
    Z_i = \frac{c_i}{\sigma_i / \sqrt{n}}
\end{equation}
where $c_i$ is the $i'th$ coefficient and $\sigma_i$ the standard deviation of the corresponding input that is multiplied with $c_i$. In traditional statistical analysis one would correct for degrees of freedom by the square root of the number of independent samples, $n$. In image analysis, when considering pixels as samples one has to be careful not confusing the number of pixels entering the analysis with independent samples. Pixels are inherently spatially correlated to high extent. In our studies millions of pixels has entered the analysis which would yield almost any coefficient significant if considered independent samples. We therefore leave out the correction for degrees of freedom and reduce Eq. \ref{eq:zscore_org} to, 
\begin{equation}\label{eq:zscore_mod}
    Z_i = \frac{c_i}{\sigma_i}
\end{equation}
This leaves out the study of significance while we can consider relative importance of the different inputs instead.

\section{Results}
\label{sec:res}
\begin{figure*}[htb]
\centering
\begin{subfigure}[b]{0.45\linewidth}
    \includegraphics[width=\linewidth]{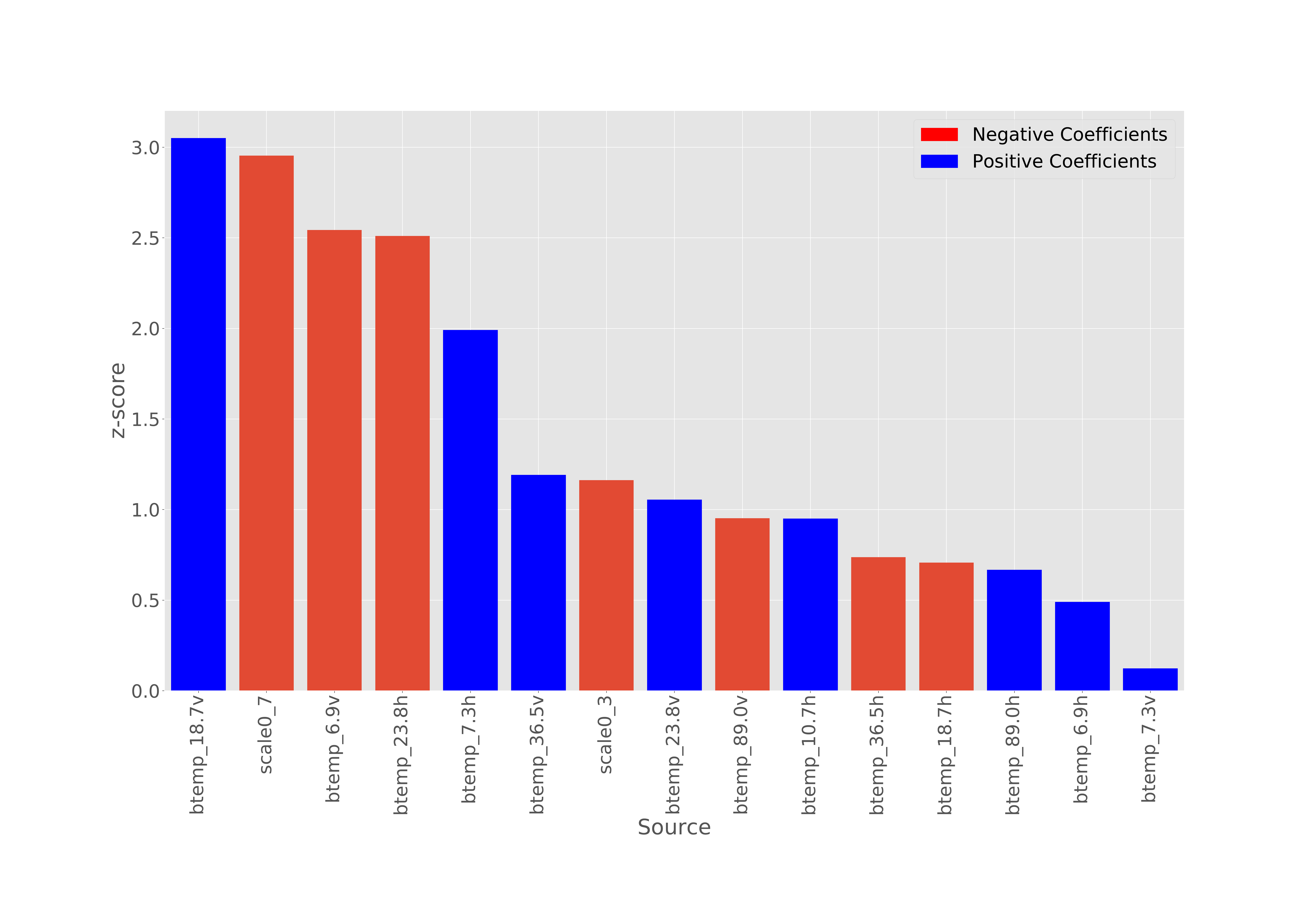}
    \caption{Small model.}
    \label{fig:all_small}
\end{subfigure}
\quad
\begin{subfigure}[b]{0.45\linewidth}
    \includegraphics[width=\linewidth]{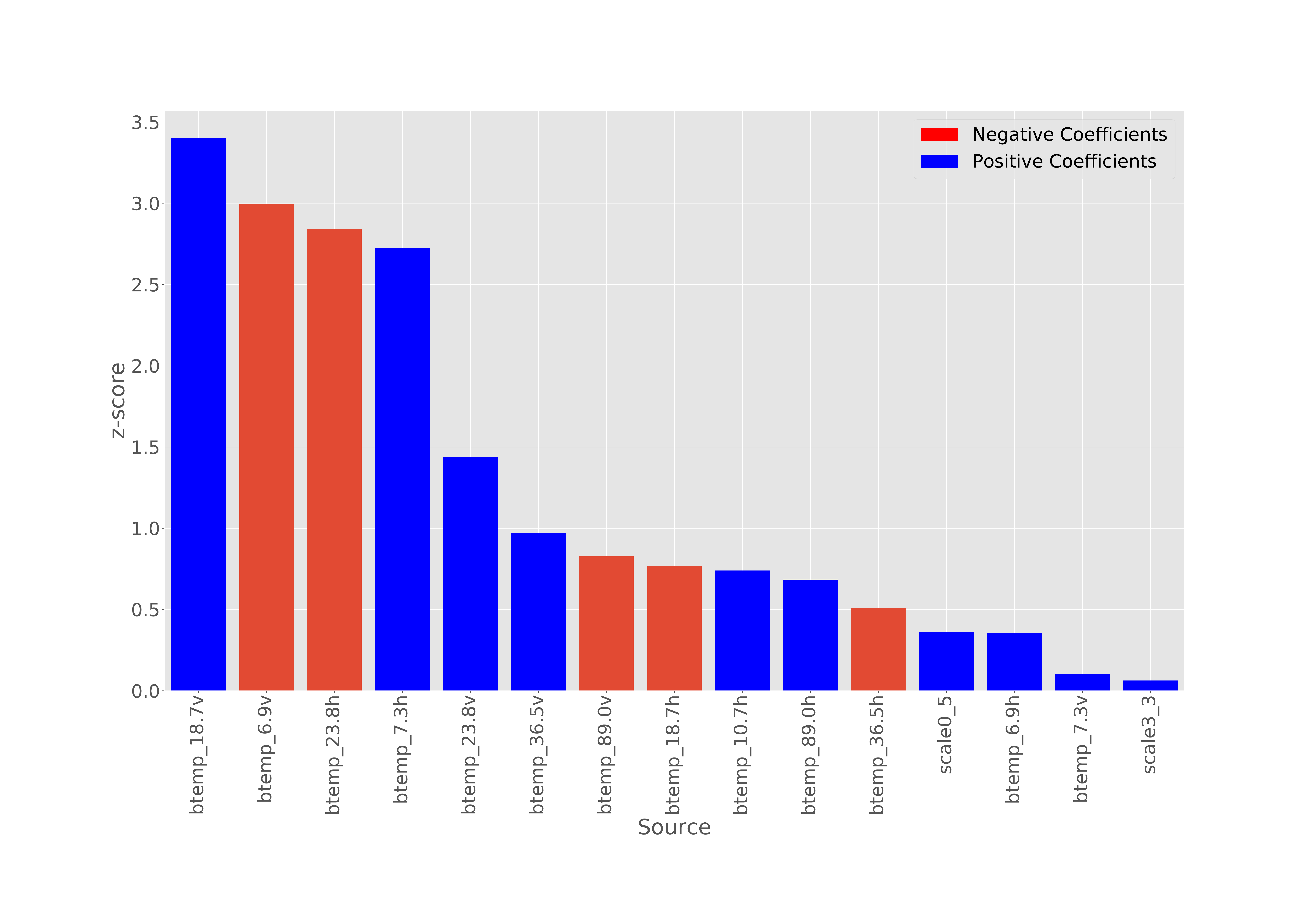}
    \caption{Large model.}
    \label{fig:all_large}
\end{subfigure}
\caption{First 14 Z-scores in descending order of importance.}\label{fig:all}
\end{figure*}

The coefficients studied in Figure \ref{fig:all_small} are from a model with 14 components in each input group of the mixing layer. By input group we mean the six different input groups, Scale-0, Scale-2, Scale-4, Scale-8, Scale-16 and AMSR2 brightness temperatures (b\_temp) as shown in Figure \ref{fig:model}. When the output layer size, i.e.\ the number of components in each group of Scale-2 to Scale-16, is increased we see a redistribution of relative importance only among the image features, Figure \ref{fig:all_large}. 

Generally, the relative importance of the AMSR2 brightness temperatures does not change with change of the model size on the image feature side. Only very few input channels change their order of importance and only slightly.



\begin{figure}[htb]
\centering
\begin{subfigure}[b]{\linewidth}
    \includegraphics[width=\linewidth]{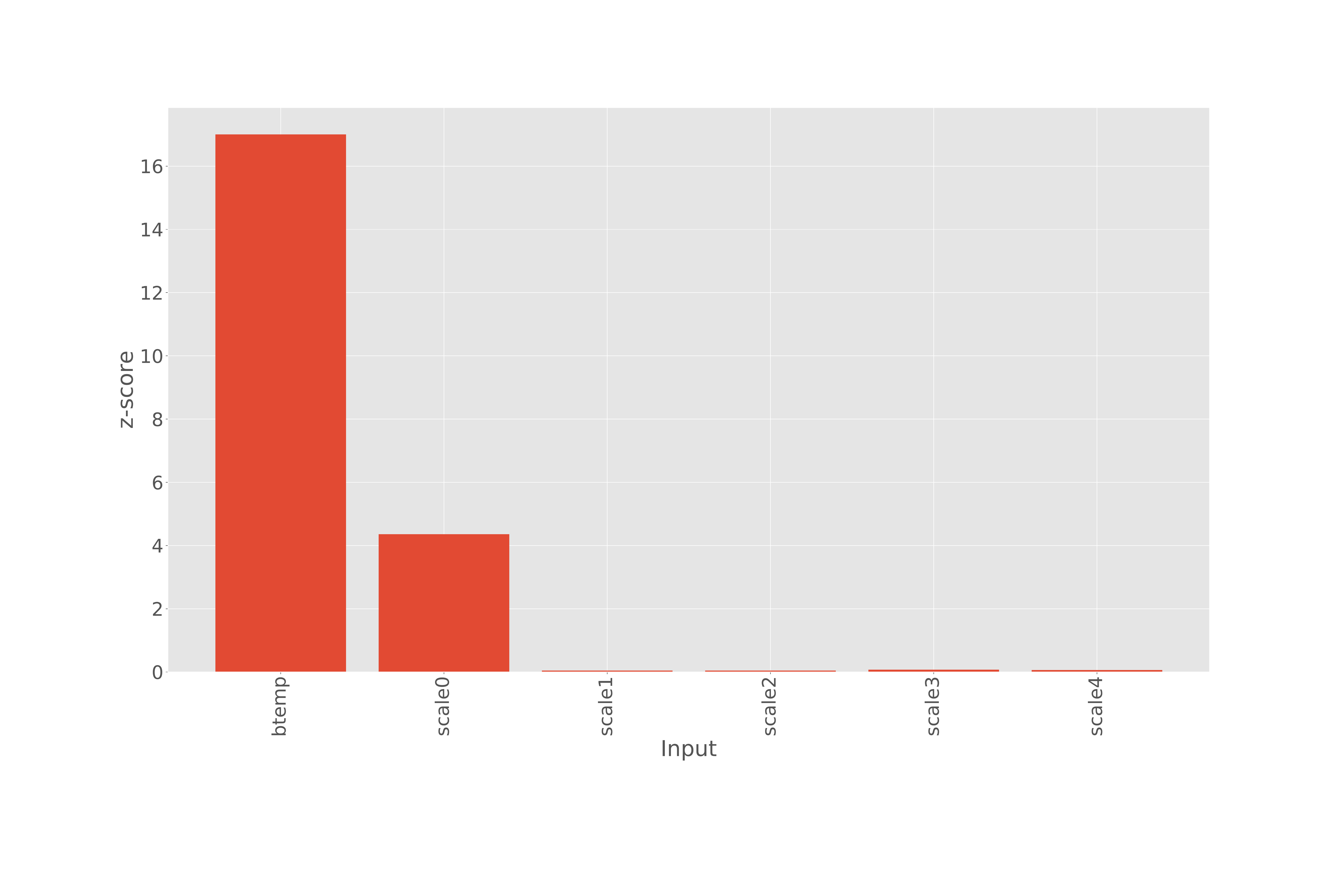}
    \caption{Small model.}\label{fig:sum_small}
\end{subfigure}

\begin{subfigure}[b]{\linewidth}
    \includegraphics[width=\linewidth]{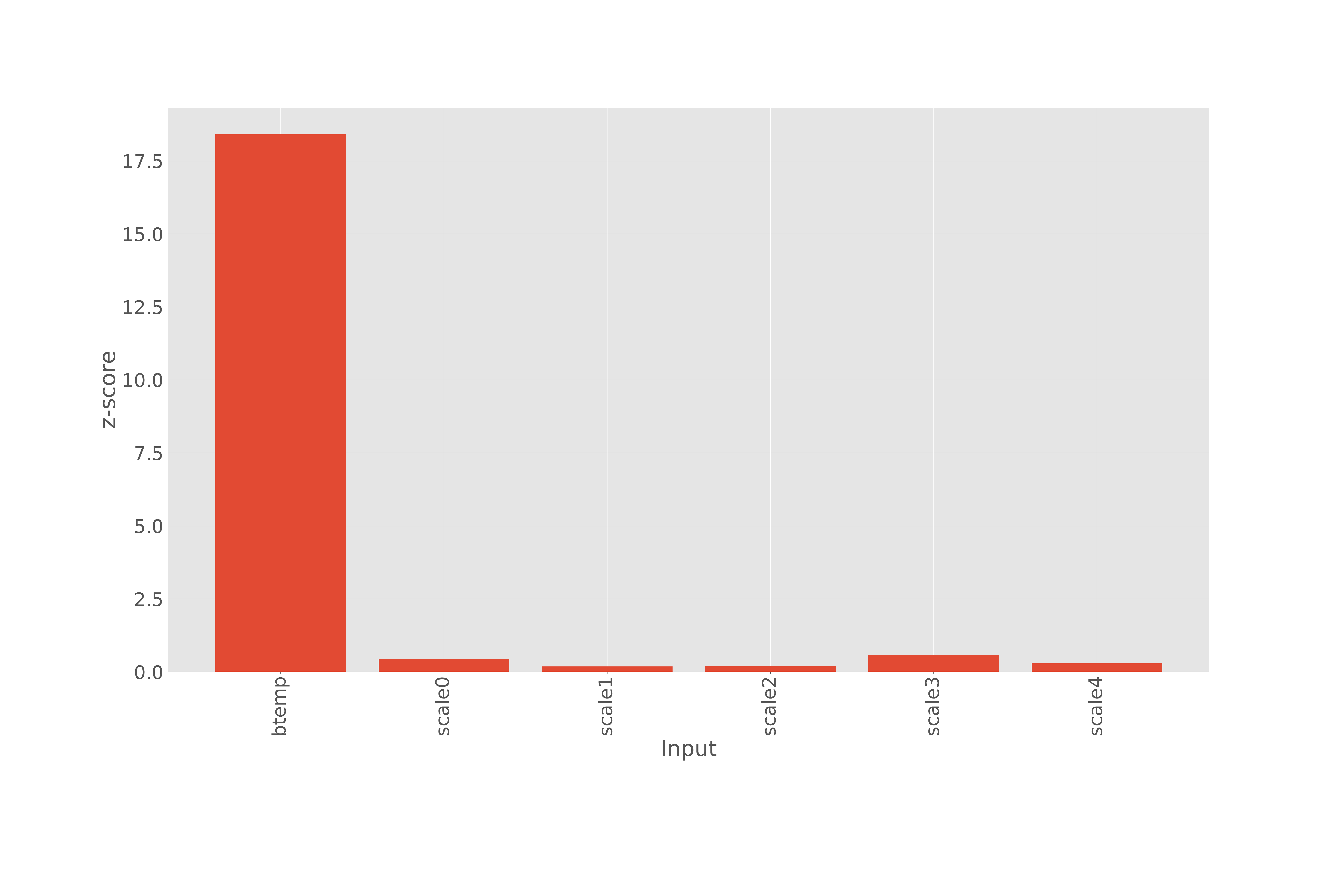}
    \caption{Large model.}\label{fig:sum_large}
\end{subfigure}
\caption{Z-scores summed over input groups, i.e. MWR (btemp), scale-0, scale-2, etc.}\label{fig:sum}
\end{figure}

In Figure \ref{fig:sum} Z-scores within each group are summed to show the overall importance of each group.

\section{Conclusion}
\label{sec:conc}
For the task of sea ice prediction, it seems that the relative importance of inputs favor the MWR data. This is not surprising as the ice charts that represent the ground truth, to a large extent consists of homogeneous areas, i.e.\ the edge pixels are relatively few compared to the total amount.

The size of the network had a small effect only on the relationship between Z-scores associated with the image features and not on the relative importance between image features and MWR inputs. The relationship between the different image feature scales change slightly, see Figures \ref{fig:sum_small} and \ref{fig:sum_large}, but the single most important image feature remains to be from Scale-0. This is interesting because it is likely to mean that the CNN uses coarse scale data from MWR to predict overall homogeneous areas. Thereby, the image features from the second CNN layer only provides edge and other higher resolution information.

It was found during the experiments that the activation function on the mixing layer affects the variance and thereby the estimation of Z-scores. The commonly used Rectified Linear Unit function lead to "dead" nodes in the Large model, i.e.\ the variances of the respective output were zero. Instead a linear activation function was used for these studies. It should also be noted that the standard deviation of AMSR2 data should be calculated prior to the upsampling, as it otherwise leads to extremely low estimates, i.e. very high z-scores. Here, the AMSR2 inputs where normalized to zero mean and unit variance prior to the model training and evaluation. 

\section{Acknowledgements}
\label{sec:page}

The authors would like thank our collaborators, at the Danish Meteorological institute, Matilde Brandt Kreiner, Jørgen Buus-Hinkler, and from the Technical Univeristy of Denmark Henning Skriver, Roberto Saldo and from Harnvig Arctic \& Maritime, Klaus Harnvig. Together in the danish research project ASIP (Automated Sea Ice Products) we have prepared the dataset and developed the models that were used for the statistical analysis of coefficients presented here. 
The authors would also like to thank the Innovation Fund Denmark for the financial support, Grant:7049-00008B, and the support of NVIDIA Corporation with the donation of the GPU used for this research.

\bibliographystyle{IEEEbib}
\bibliography{refs}

\end{document}